\documentclass[11pt,a4paper]{article}
\usepackage[hyperref]{acl2020}
\usepackage{times}
\usepackage{latexsym}
\usepackage{adjustbox}
\usepackage{amsmath}
\usepackage{amssymb}
\usepackage{booktabs}
\usepackage{dirtytalk}
\usepackage{enumitem}
\usepackage{microtype}

\usepackage{siunitx}
\newcommand{\term}[1]{\textbf{#1}}
\usepackage{longtable}
\usepackage{subcaption}
\usepackage{xfrac}
\usepackage{supertabular}

\usepackage{tipa}
\usepackage{multirow}
\usepackage{tipa}
\usepackage{graphicx} 
\graphicspath{{./figs/}} 

\def\Snospace~{\S{}}

\DeclareMathAlphabet{\mathcal}{OMS}{cmsy}{m}{n}
\DeclareMathAlphabet{\mathbb}{U}{msb}{m}{n}
\DeclareMathOperator{\NMI}{NMI}

\newcommand\semantics{\ensuremath{\mathbf{v}_i}}
\newcommand\wordform{\ensuremath{\mathbf{w}_i}}

\newcommand\class{\ensuremath{c_i}}

\newcommand{\semanticspace}{\ensuremath{\mathbb{R}^d}}
\newcommand{\alphabet}{\ensuremath{\Sigma}}
\newcommand{\wordformspace}{\ensuremath{\alphabet^*}}
\newcommand{\classspace}{\ensuremath{\mathcal{C}}}

\usepackage{todonotes}

\aclfinalcopy 

\DeclareMathOperator{\Entr}{H}
\DeclareMathOperator{\MI}{MI}
\DeclareMathOperator{\U}{U}

\DeclareMathOperator{\Unc}{U}

\renewcommand{\Pr}{p}
\newcommand{\defn}[1]{\textbf{#1}}
\DeclareMathOperator{\calD}{\mathcal{D}}

\everypar{\looseness=-1}

\title{Predicting Declension Class from Form and Meaning} 

\newcommand{\ucambridge}{\textnormal{\text{\textipa{D}}}}
\newcommand{\ethz}{\textnormal{\text{\textipa{Q}}}}
\newcommand{\fairesearch}{\textnormal{\text{\textipa{@}}}}
\newcommand{\nyu}{\textnormal{\textipa{\textsch}}}
\newcommand{\jhu}{\textnormal{\text{\textipa{Z}}}}
\newcommand{\york}{\textnormal{\text{\textipa{Y}}}}

\author{Adina Williams$^\fairesearch$ Tiago Pimentel$^\ucambridge$ Arya D. McCarthy$^\jhu$\\
\bf Hagen Blix$^\nyu$ Eleanor Chodroff$^\york$ Ryan Cotterell$^{\ucambridge,\ethz}$ \\
  $^\fairesearch$Facebook AI Research~\;~$^\ucambridge$University of Cambridge~\;~$^\jhu$Johns Hopkins University~\ \\
  ~$^\nyu$New York University~\;~$^\york$University of York~\;~$^\ethz$ETH Z\"{u}rich \\
  \texttt{adinawilliams@fb.com},~\;~ \texttt{tp472@cam.ac.uk},~\\  \texttt{arya@jhu.edu},~\;~
  \texttt{hagen.blix@nyu.edu},~\\ 
  \texttt{eleanor.chodroff@york.ac.uk},\;~ \texttt{ryan.cotterell@inf.ethz.ch}
}

\begin{document}
\maketitle

\begin{abstract}
The noun lexica of many natural languages are divided into several declension classes with characteristic morphological properties. Class membership is far from deterministic, but the phonological form of a noun and its meaning can often provide imperfect clues. Here, we investigate the strength of those clues. More specifically, we operationalize ``strength'' as measuring how much information, in bits, we can glean about declension class from knowing the form and meaning of nouns. We know that form and meaning are often also indicative of grammatical gender---which, as we quantitatively verify, can itself share information with declension class---so we also control for gender. We find for two Indo-European languages (Czech and German) that form and meaning share a significant amount of information with class (and contribute additional information beyond gender). The three-way interaction between class, form, and meaning (given gender) is also significant.
Our study is important for two reasons: First, we introduce a new method that provides additional quantitative support for a classic linguistic finding that form and meaning are relevant for the classification of nouns into declensions. Second, we show not only that individual declension classes vary in the strength of their clues within a language, but also that the variations between classes vary \emph{across languages}. The code is publicly available at \url{https://github.com/rycolab/declension-mi}.
\end{abstract}

\section{Introduction}

\begin{figure}[ht]
    \centering
    \includegraphics[width=\columnwidth]{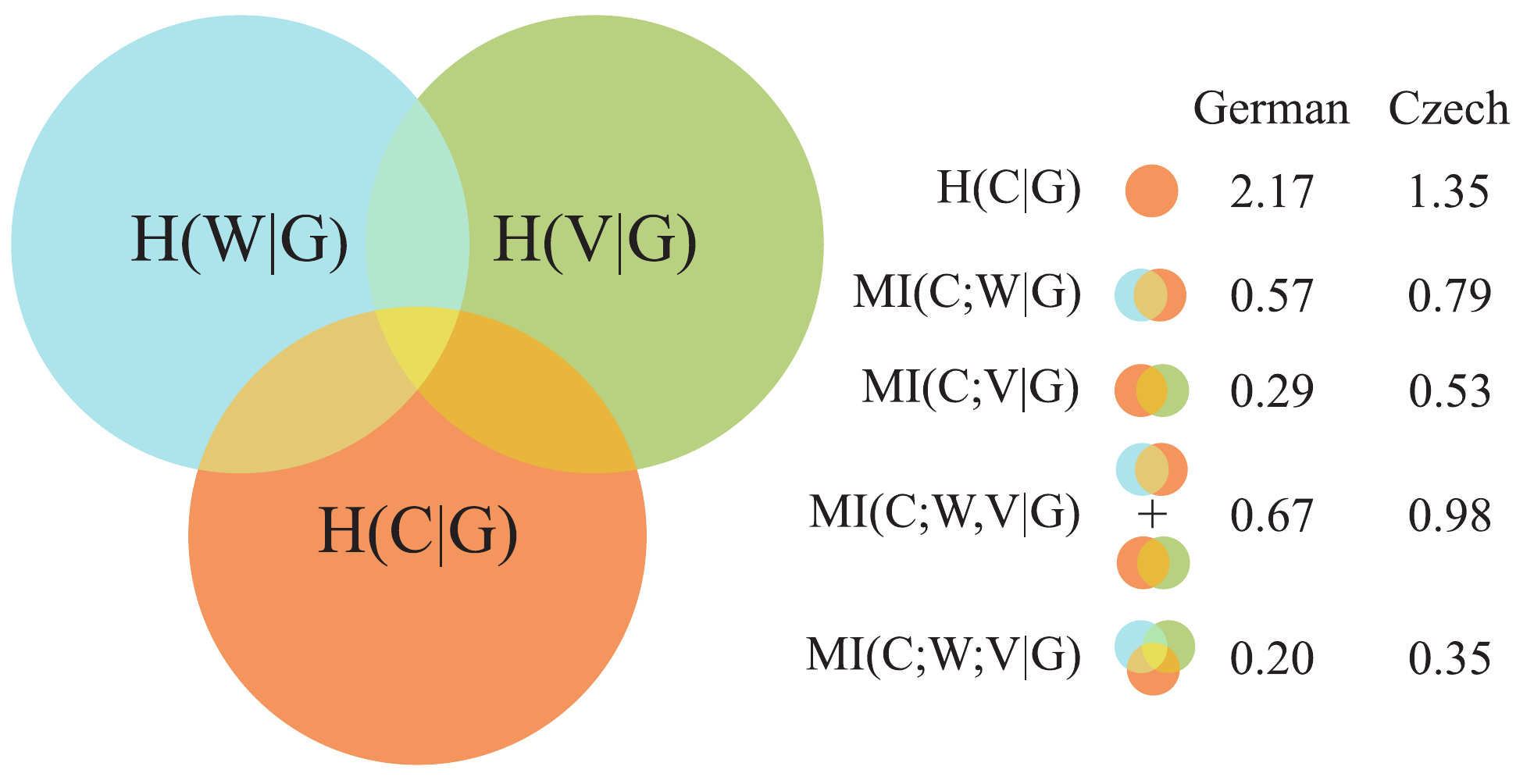} 
    \caption{The conditional entropies ($\Entr$) and mutual information quantities ($\MI$) of form ($W$), meaning ($V$), and declension class ($C$), given gender ($G$) in German and Czech.}
    \label{fig:declension_mi}
\end{figure}

To an English speaker learning German, it may come as a surprise that one cannot necessarily predict the plural form of a noun from its singular. 
This is because pluralizing nouns in English is relatively simple: Usually we merely add an \textit{-s} to the end (e.g., \textit{cat} \(\mapsto\) \textit{cat\textbf{s}}). 
Of course, not all English nouns follow such a simple rule (e.g., \textit{child} \(\mapsto\) \textit{child\textbf{ren}}, \textit{sheep} \(\mapsto\) \textit{sheep}, etc.), but those that do not are few in number. Compared to English, German has comparatively many common morphological rules for inflecting nouns. For example, some plurals are formed by adding a suffix to the singular: \textit{Insekt} `insect' \(\mapsto\) \textit{Insekt-\textbf{en}}, \textit{Hund} `dog' \(\mapsto\) \textit{Hund-\textbf{e}}, \textit{Radio} `radio' \(\mapsto\) \textit{Radio-\textbf{s}}. For others, the plural is formed by changing a stem vowel:\footnote{This vowel change, \term{umlaut}, corresponds to fronting.}\ \textit{Mutter} `mother' \(\mapsto\) \textit{M\textbf{\"{u}}tter}, or \textit{Nagel} `nail' \(\mapsto\) \textit{N\textbf{\"{a}}gel}. Some others form plurals with both suffixation and vowel change: \textit{Haus} `house'  \(\mapsto\) \textit{H\textbf{\"a}us-er} and \textit{Koch} `chef'  \(\mapsto\) \textit{K\textbf{\"o}ch-e}. Still others, like \textit{Esel} `donkey', have the same form in plural and singular. The problem only worsens when we consider other inflectional morphology, such as case.

Disparate plural formation and case rules of the kind described above split nouns into \textbf{declension classes}. To know a noun's declension class is to know which morphological form it takes in which context (e.g., \citealt{benveniste1935, wurzel1989, nubling2008, ackerman2009, ackerman2013,
beniamine2016, bonami2016}). But, this begs the question: What clues can we use to predict the class for a noun? In some languages, predicting declension class is argued to be easier if we know the noun's phonological form \cite{aronoff1992, dressler1996} or lexical semantics \cite{carstairs1994, corbett2000}. However, semantic and phonological clues are, at best, only very imperfect hints as to class \cite{wurzel1989, harris1991, harris1992,  aronoff1992, halle1994, corbett2000, aronoff2007}.
Given this, we quantify \emph{how much} information a noun's form and meaning  share with its class, and determine whether that amount of information is uniform across classes.

To do this, we measure the \textbf{mutual information} \citep{cover2012elements} both between declension class and meaning (i.e.,\ distributional semantic vector) and between declension class and form (i.e.,\ orthographic form), as in \autoref{fig:declension_mi}. We select two Indo-European languages (Czech and German) that have declension classes. We find that form and meaning both share significant amounts of information, in bits, with declension class in both languages. We further find that form clues are stronger than meaning clues; for form, we uncover a relatively large effect of 0.5--0.8 
bits, while, for lexical semantics, a moderate one of 0.3--0.5 bits.  We also measure the three-way interaction between form, meaning, and class, finding that phonology and semantics contribute overlapping information about class. Finally, we analyze individual inflection classes and uncover that the amount of information they share with form and meaning is not uniform across classes or languages. 

\section{Declension Classes in Language}

The morphological behavior of declension classes is quite complex. Although various factors are undoubtedly relevant, we focus on phonological and lexical semantic ones here. We have ample reason to suspect that phonological factors might affect class predictability. In the most basic sense, the form of inflectional suffixes are often altered based on the identity of the final segment of the stem. For example, the English plural suffix is spelled as \textit{-s} after most consonants, like in \textit{cat\textbf{s}}, but as \textit{-es} if it appears after an \textit{s, sh, z, ch} etc., like in `\textit{moss\textbf{es}}', `\textit{rush\textbf{es}}', `\textit{quizz\textbf{es}}', `\textit{beach\textbf{es}}' etc. Often differences such as these in the spelling of plural affixes or declension class affixes are due to phonological rules that are noisily realized in orthography; there could also be regularities between form and class that do not correspond to phonological rules but still have an effect. For example, statistical regularities over phonological segments in continuous speech guide first-language acquisition \citep{maye2002}, even over non-adjacent segments \citep{newport2004}. Statistical relationships have also been uncovered between the sounds in a word and the word's syntactic category \citep{farmer2006, monaghan2007, sharpe2017} and between the orthographic form of a word and its argument structure valence \citep{williams2018}. Thus, we expect the form of a noun to provide clues to declension class.  

Semantic factors too are often relevant for determining certain types of morphologically relevant classes, such as grammatical gender, which is known to be related to declension class. It has been claimed that there are only two types of gender systems: \defn{semantic systems} (where only semantic information is required) and \defn{formal systems} (where semantic information as well as morphological and phonological factors are relevant) \citep[294]{corbett2000}. 
Moreover, a large typological survey, \newcite{qian2016}, finds that meaning-sensitive grammatical properties, such as gender and animacy, can be decoded well from distributional word representations for some languages, but less well for others. These examples suggest that it is worth investigating whether noun semantics provides clues about declension class. 

Lastly, form and meaning might interact with one another, as in the case of \textbf{phonaesthemes} where the sounds of words provide non-arbitrary clues about their meanings \citep{sapir1929,wertheimer1958, holland1964,maurer2006,monaghan2014, donofrio2014, dingemanse2015, dingemanse2018, pimentel2019}. Therefore, we check whether form and meaning together share information with declension class.

\subsection{Orthography as a proxy for phonology?}\label{subsec:orthography}

We motivate an investigation into the relationship between the form of a word and its declension class by appealing, at least partly, to phonological motivations. However, we make the simplifying assumption that phonological information is adequately captured by orthographic word forms---i.e.,\ strings of written symbols, which are also known as \textbf{graphemes}. In general, one should question this assumption \citep{vachek1945,luelsdorff1987, sproat2000, sproat2012, neef2012}. For the particular languages we investigate here---Czech and German---it is less problematic, as they are have fairly ``transparent'' mappings between spelling and pronunciation \citep{matvejvcek1998, miles2000, caravolas2001}, which enables them to achieve higher performance on grapheme-to-phoneme conversion than do English and other ``opaque'' orthographic systems \citep{schlippe2012}. These studies suggest that we are justified in taking orthography as a proxy for phonological form. Nonetheless, to mitigate against any phonological information being inaccurately represented in the orthographic form (e.g., vowel lengthening in German), several of our authors, who are fluent reader--annotators of our languages, checked our classes for any unexpected phonological variations. We exhibit examples in \autoref{sec:Data}. 

\subsection{Distributional Lexical Semantics}

We adopt a distributional approach to lexical semantics (\citealt{harris1954,mitchell2010,turney2010,bernardi-etal-2015-distributional,clark2015}; \textit{inter alia}) that relies on pretrained word embeddings for this paper. We do this for multiple reasons: First, distributional semantic approaches to create word vectors, such as \textsc{word2vec} \citep{mikolov2013}, have been shown to do well at extracting lexical features such as animacy and taxonomic information \citep{rubinstein2015} and can also recognize semantic anomaly \citep{vecchi2011linear}. Second, the distributional approach to lexical meaning yields a straightforward procedure for extracting ``meaning'' from text corpora at scale.

\subsection{Controlling for grammatical gender?}\label{subsec:genderclass}

Grammatical gender has been found to interact with lexical semantics \citep{schwichtenberg2004, williams2019, lera}, and often can be determined from form \citep{brooks1993, dobrin1998, frigo1998, starreveld2004}. This means that it cannot be ignored in the present study.
While the precise nature of the relationship between declension class and gender is far from clear, it is well established that the two should be distinguished (\citealt{aronoff1992, Wiese.2000, Kurschner.Nubling.2011}; \textit{inter alia}). 
We first measure the amount of information shared between gender and class, according to the methods described in \S 4, to verify that the predicted relationship exists. We then verify that gender and class overlap in information in German and Czech to a high degree, but that we cannot reduce one to the other (see \autoref{tab:results-gender} and \autoref{results}). We proceed to control for gender, and subsequently measure how much \emph{additional} information form and meaning provide about declension class.

\section{Data}\label{sec:Data}

\begin{table*}
\setlength{\belowcaptionskip}{-10pt}
\begin{center}
\small
\begin{tabular}{lcccccccc}
\toprule
 &  Original &  Final &  Training & Validation & Test & Average Length & \# Classes \\ \midrule
\bf Czech  & 3011 & 2672 & 2138 & 267 & 267 & 6.26 & 13\\
\bf German & 4216 & 3684 & 2948 & 368 & 368 & 5.87 & 16\\ \bottomrule
\end{tabular}
\end{center}
\caption{Number of words in dataset. Counts per language-category pair are listed both before and after preprocessing, train-validation-test split, average stem length, and \# of classes. Since we use 10-fold cross-validation, all instances are included in the test set at some point, and are used to estimate the cross-entropies in \autoref{sec:variational}.} \label{tab:data-split}
\end{table*}

For our study, we need orthographic forms of nouns, their associated word vectors, and their declension classes.  Orthographic forms can be found in any large text corpus or dictionary. We isolate noun \term{lexemes} (i.e., or syntactic category--specific representations of words) by language. We select Czech nouns from UniMorph \citep{kirov-etal-2018-unimorph} and German nouns from CELEX2 \cite{celex2}. For lexical semantics, we trained 300-dimensional \textsc{word2vec} vectors on language-specific Wikipedia.\footnote{We use the \textsc{gensim} toolkit \citep{gensim}.}

We select the nominative singular form as the donor for both orthographic and lexical semantic representations because it is the \textbf{lemma} in Czech and German. It is also usually the \textbf{stem} for the rest of the morphological paradigm. We restrict our investigation to monomorphemic lexemes because: (i) one stem can take several affixes which would multiply its contribution to the results, and (ii) certain affixes come with their own class.\footnote{Since these require special treatment, they are set aside.}

Compared to form and meaning, declension class is a bit harder to come by, because it requires linguistic annotation. We associated lexemes with their classes on a by-language basis by relying on annotations from fluent speaker--linguists, either for class determination (for Czech) or for verifying existing dictionary information (for German). For Czech, declension classes were derived by an edit distance heuristic over affix forms, which grouped lemmata into subclasses if they received the same inflectional affixes (i.e.,\ they constituted a morphological paradigm). If orthographic differences between two sets of suffixes in the lemma form could be accounted for by positing a phonological rule, then the two sets were collapsed into a single set; for example, in the ``feminine \textit{-a}'' declension class, we collapsed forms for which the dative singular suffix surfaces as \textit{-e} following a coronal continuant consonant (\textit{figurka}:\textit{figur\textbf{c}e} `figurine.\textsc{dat.sg}'), \textit{-i} following a palatal nasal (\textit{pira\u{n}a}:\textit{pira\textbf{\u{n}}i} `piranha.\textsc{dat.sg}'), and as \textit{-\u{e}} following all other consonants (\textit{kr\'ava}:\textit{kr\'a\textbf{v}\u{e}} `cow.\textsc{dat.sg}').  As for meaning, descriptively, gender is roughly a superset of declension classes in Czech; among the masculine classes, animacy is a critical semantic feature, whereas form seems to matter more for feminine and neuter classes. 
 
For German, nouns came morphologically parsed and lemmatized, as well as coded for class in CELEX2. We also use CELEX2 to isolate monomorphemic noun lexemes and bin them into classes; however, CELEX2 declension classes are more fine-grained than traditional descriptions of declension class---mappings between CELEX2 classes and traditional linguistic descriptions of declension class \citep{alexiadou2008} are provided in \autoref{tab:NounClass} in the Appendix. The CELEX2 declension class identifier scheme has multiple subparts. Each declension class identifier includes: (i) the number prefix (being `S' is for singular, or `P' for plural), (ii) the morphological form identifier---zero refers to paradigmatically missing forms (e.g.,\ plural is zero for \defn{singularia tantum} nouns), and other numbers refer to a form identifier of particular morphological processes (e.g.,\ genitive applies an additional suffix for singular masculine nouns, but never for feminines)---and (iii) an optional `u' identifier, which refers to vowel umlaut, if present. More details of the German preprocessing steps are in the Appendix.

After associating nouns with forms, meanings, and classes, we perform exclusions: Because frequency affects class entropy \citep{parker2015}, we removed all classes with fewer than 20 lexemes.\footnote{We ran another version of our models that included all the original classes and observed no notable differences.}\  We subsequently removed all lexemes which did not appear in our \textsc{word2vec} models trained on Wikipedia dumps. The final tally of Czech yields 2672 nouns in 13 declension classes, and the final tally of German yields 3684 nouns in 16 declension classes, which can be broken into 3 types of singular and 7 types of plural. \autoref{tab:results5-meaning} in the Appendix provides final lexeme counts by declension class.

The remaining lexemes were split into 10 folds: one for testing, another for validation, and the remaining eight for training. \autoref{tab:data-split} shows train--validation--test splits, average length of nouns, and number of declension classes, by language. 

\section{Methods}\label{sec:methods}

\paragraph{Notation.}
We define each lexeme in a language as a triple. Specifically, the $i^{\text{th}}$ triple consists of an orthographic word form $\wordform$, a distributional semantic vector $\semantics$ that encodes the lexeme's semantics, and a declension class $\class$. We assume these triples follow a (unknown) probability distribution $p(\mathbf{w}, \mathbf{v}, c)$---which can be marginalized to obtain $p(c)$, for example. We take the space of word forms to be the Kleene closure over a language's alphabet $\alphabet$; thus, we have $\wordform \in \wordformspace$.  Our distributional semantic space is a high-dimensional real vector space $\semanticspace$ where $\semantics \in \semanticspace$. The space of declension classes is language-specific and contains as many elements as the language has classes, i.e., $\classspace = \{1, \ldots, K\}$  where $\class \in \classspace$. For each noun, a gender $g_i$ from a language-specific space of genders $\mathcal{G}$ is associated with the lexeme. In both Czech and German, $\mathcal{G}$ contains three genders: feminine, masculine, and neuter. We also consider four random variables: a $\Sigma^*$-valued random variable $W$, an $\mathbb{R}^d$-valued random variable $V$, a $\mathcal{C}$-valued random variable $C$ and a $\mathcal{G}$-valued random variable $G$.

\paragraph{Bipartite Mutual Information.}
Bipartite $\MI$ (or, simply $\MI$) is a symmetric quantity that measures how much information (in bits)
two random variables share. In the case of $C$ (declension class) and $W$ (orthographic form), we have
  \begin{align}\label{eq:MI}
    \MI(C; W) &= \Entr(C) - \Entr(C \mid W) 
  \end{align}%
As can be seen, $\MI$ is the difference between an unconditional and a conditional entropy. The unconditional entropy is defined as
\begin{align} \label{eq:entropy}
    \Entr(C) &=-\sum_{c \in \mathcal{C}} \Pr(c) \log \Pr(c) 
\end{align}
and the conditional entropy is defined as 
\begin{align} \label{eq:entropy_conditional}
    \Entr(C \mid W) &= \\
    &-\sum_{c \in \classspace} \sum_{\mathbf{w} \in \Sigma^{*}} \Pr(c, \mathbf{w}) \log \Pr(c \mid \mathbf{w}) \nonumber
\end{align}
The mutual linformation $\MI(C; W)$ naturally encodes
how much the orthographic word form tells us about its corresponding
lexeme's declension class. Likewise, to measure the interaction between declension class and lexical semantics, we also consider the bipartite mutual information $\MI(C; V)$. 

\paragraph{Tripartite Mutual Information.}
To consider the interaction between three random variables at once,
we need to generalize $\MI$ to three classes. One can calculate tripartite $\MI$ as follows:
\begin{align} \label{eq:threewayMI}
\MI(C; W; V) &= \\
&\MI(C ; W) - \MI(C; W \mid V) \nonumber
\end{align}
As can be seen, tripartite $\MI$ is the difference between
a bipartite $\MI$ and a conditional bipartite MI.
The conditional bipartite $\MI$ is defined as
\begin{align}
\MI(C; W \mid V) &= \Entr(C \mid V) - \Entr(C \mid W,  V)
\end{align}%
Essentially, \autoref{eq:threewayMI} is the difference between how much $C$ and $W$ interact and how much they interact after ``controlling'' for the meaning $V$.%
\footnote{We emphasize here the subtle, but important, typographic distinction between $\MI(C; W; V)$ and $\MI(C; W, V)$. (The difference in notation lies in the comma replacing the semicolon.) While the first (tripartite MI) measures the amount of (redundant) information shared by the three variables, the second (bipartite) measures the (total) information that class shares with either the form \emph{or} the lexical semantics.}

\paragraph{Controlling for Gender.}
Working with mutual information also gives
us a natural way to control for quantities
that we know influence meaning and form. We
do this by considering conditional MI. We consider
both bipartite and tripartite conditional mutual information.
These are defined as follows:
\begin{subequations} \label{eq:MI_conditional}%
\begin{align}
    \MI(C; W \mid &G ) = \\
    &\Entr(C \mid G) - \Entr(C \mid W, G) \nonumber \\
    \MI(C; W; V \mid &G) = \\
    & \MI(C ; W \mid G) - \MI(C; W \mid V, G) \nonumber
\end{align}
\end{subequations}
Estimating these quantities tells us how much $C$ and $W$ (and, in the case of tripartite MI, $V$ also)
interact after we take $G$ (the grammatical gender) out of the picture. \autoref{fig:declension_mi} provides a graphical summary for this section until this point.

\paragraph{Normalization.}
To further contextualize our results, we 
consider two normalization schemes for MI. Normalizing renders $\MI$ estimates across languages more directly comparable \citep{gates2019}.
We consider the \textbf{normalized mutual information}, i.e., which \emph{fraction} of the unconditional entropy is the mutual information:
\begin{equation}
    \NMI(C; W) = \frac{\MI(C; W)}{\min\{\Entr(C), \Entr(W)\}} 
\end{equation}%
This yields a percentage of the entropy that the mutual information accounts for---a more interpretable notion of the predictability between class and form or meaning.  
In practice, $\Entr(C) \ll \Entr(W)$ in most cases and our normalized
mutual information is termed the \term{uncertainty coefficient} \citep{theil1970}:
\begin{equation}
    \Unc(C \mid W) = \frac{\MI(C; W)}{\Entr(C)}
\end{equation}%

\section{Computation and Approximation}\label{sec:variational}
In order to estimate the mutual information quantities of
interest per \autoref{sec:methods}, we need
to estimate a variety of entropies. We derive
our mutual information estimates from a corpus $\mathcal{D} = \{(\mathbf{v}_i, \mathbf{w}_i, c_i)\}_{i=1}^N$. 

\subsection{Plug-in Estmation of Entropy}
\label{sec:estimate_entropy}
The most straightforward quantity to estimate is $\Entr(C)$. Given a corpus, we may use plug-in estimation: We compute
the empirical distribution over declension classes from $\mathcal{D}$. 
Then, we plug that empirical distribution over declension classes $\mathcal{C}$ into
the formula for entropy in \autoref{eq:entropy}. 
This estimator is biased \cite{paninski}, but is a suitable choice given that we have only a few declension classes and a large amount of data. Future work will explore whether choice of estimator \citep{miller1955,hutter2001,archer2013, archer2014} could affect
the conclusions of studies such as this one.

\subsection{Model-based Estimation of Entropy} \label{sec:estimate_entropy_cond}
In contrast, estimating $\Entr(C \mid W)$ is non-trivial.
We cannot simply apply plug-in estimation because we cannot compute the infinite sum over $\Sigma^*$ that is required. Instead, we follow previous work \cite{brown-etal-1992-estimate,pimentel2019} in using 
the cross-entropy upper bound to approximate $\Entr{}(C \mid W)$ with a 
model. More formally, for any probability distribution $q(c \mid \mathbf{w})$, we have
\begin{align}
\Entr(C \mid W) &\leq \Entr_q(C \mid W) \label{eq:cross-entropy}\\
&= -\sum_{c \in \mathcal{C}} \sum_{\mathbf{w} \in \Sigma^*} p(c, \mathbf{w}) \log q(c \mid \mathbf{w}) \nonumber
  \end{align}
To circumvent the need for infinite sums, we use
a held-out sample $\tilde{\mathcal{D}} = \{(\tilde{\mathbf{v}}_i, \tilde{\mathbf{w}}_i, \tilde{c}_i)\}_{i=1}^M$ disjoint from $\calD$
to approximate the true cross-entropy $\Entr_q(C \mid W)$ with the following quantity 
\begin{equation}\label{eq:H_conditional}
  \hat{\Entr}_q(C \mid W) = -\frac{1}{M} \sum_{i=1}^M \log q\left(\tilde{\class{}} \mid \tilde{\wordform{}}\right)
\end{equation}
where we assume the held-out data is distributed according to the true distribution $p$. We note that $\hat{\Entr}_q(C \mid W) \rightarrow \Entr_q(C \mid W)$ as $M \rightarrow \infty$. While the exposition above focuses on learning
a distribution $q(c \mid \mathbf{w})$ for classes and forms to approximate $\Entr(C \mid W)$, the same methodology can be used to estimate all necessary conditional
entropies.

\paragraph{Form and gender: $q(c \mid \mathbf{w}, g)$.} We train one LSTM classifier \citep{hochreiter1997} for each language. The last hidden state of the LSTM models is fed into a linear layer and then a softmax non-linearity to obtain probability distributions over declension classes. To condition our model on gender, we embed
each gender and feed it into each LSTM's initial hidden state. 

\paragraph{Meaning and gender: $q(c \mid \mathbf{v}, g)$.}
We trained a simple multilayer perceptron (MLP) classifier to predict the declension class from the \textsc{word2vec} representation. When conditioning on gender, we again embed each gender class, concatenating these embeddings with the \textsc{word2vec} ones before feeding the result into the MLP. 

\paragraph{Form, meaning, and gender: $q(c \mid \mathbf{w}, \mathbf{v}, g)$.}
We again trained two LSTM classifiers, but this time, also conditioned on meaning (i.e., \textsc{word2vec}). Before training, we reduce the dimensionality of the \textsc{word2vec} embeddings from 300 to $k$ dimensions by running PCA on each language's embeddings. We then linearly transformed them to match the hidden size of the LSTMs, and fed them in. To also condition on gender, we followed the same procedures, but used half of each LSTM's initial hidden state for each vector (i.e., \textsc{word2vec} and one-hot gender embeddings).  

\begin{table*}
\begin{center}
\begin{adjustbox}{width=\linewidth}
\begin{tabular}{lcccccccccccc}
\toprule
& \multicolumn{4}{c}{\bf Form \& Declension Class (LSTM)} & \multicolumn{4}{c}{\bf Meaning \& Declension Class (MLP)}\\
 & $\Entr(C \mid G)$ & $\Entr_{Q}(C \mid W, G)$ & $\MI(C; W \mid G)$ & $\U(C \mid W, G)$  & $\Entr(C \mid G)$ & $\Entr_{Q}(C \mid V, G)$ & $\MI(C; V \mid G)$ & $\U(C \mid V, G)$ \\ \cmidrule(l){1-5} \cmidrule(l){6-9}
Czech  & 1.35  & 0.56 & \bf 0.79 & 58.8\% & 1.35  & 0.82 & \bf 0.53 & 39.4\%  \\ 
German &  2.17 & 1.60 & \bf 0.57 & 26.4\% & 2.17  & 1.88 & \bf 0.29 & 13.6\% \\  \midrule
& \multicolumn{4}{c}{\bf Both (Form and Meaning) \& Declension Class} & \multicolumn{4}{c}{\bf Tripartite $\MI$ (LSTM)}\\
 & $\Entr(C \mid G)$ & $\Entr_Q(C \mid W, V, G)$ & $\MI(C; W, V \mid G)$ & $\U(C \mid W, V, G)$ & $\MI(C; W \mid G)$ &  $\MI(C; W \mid V, G)$  & $\MI(C; W; V \mid G)$ & $\U(C \mid W ; V, G)$ \\ \cmidrule(l){1-5} \cmidrule(l){6-9}
Czech  & 1.35  & 0.37 & \bf 0.98 & 72.6\% & 0.79 & 0.44& \bf 0.35  & 25.9\%  \\ 
German & 2.17 & 1.50 & \bf 0.67 & 30.8\% & 0.57 & 0.37 & \bf 0.20 & \phantom{0}9.2\%  \\  \bottomrule
\end{tabular}
\end{adjustbox}
\end{center}
\caption{MI between form and class (top-left), meaning and class (top-right), both form and meaning and class (bottom-left), and tripartite $\MI$ (bottom-right). All values are calculated given gender, and bold if significant.} \label{tab:results-form}
\end{table*}

\paragraph{Optimization.}
We trained all classifiers using Adam \citep{kingma2015adam} and the code was implemented using PyTorch. Hyperparameters---number of training epochs, hidden sizes, PCA compression dimension ($k$), and number of layers---were optimized using Bayesian optimization with a Gaussian process prior \citep{snoek2012practical}. We explore a maximum of 50 models for each experiment, maximizing the expected improvement on the validation set.

\subsection{An Empirical Lower Bound on MI}
With our empirical approximations of the desired entropy measures, we can calculate the desired approximated $\MI$ values, e.g.,
\begin{align}
    \MI(C ; W \mid G) &\approx \\
    &\hat{\Entr}(C \mid G) - \hat{\Entr}_q(C \mid W, G) \nonumber
\end{align}
where $\hat{\Entr}(C \mid G)$ is the plug-in estimation of the entropy.
Such an approximation, though, is not ideal, since we do not know if the true $\MI$ is approximated by above or below. Since we use a plug-in estimator for $\hat{\Entr}(C \mid G)$, which \emph{underestimates} entropy, and since $\Entr_q(C \mid W, G)$ is estimated with a cross-entropy \emph{upperbound}, we have
\begin{align*}
    \MI(C ; W \mid G) &=  \Entr(C \mid G) - \Entr(C \mid W, G) \\
    &\gtrapprox \hat{\Entr}(C \mid G) - \Entr(C \mid W, G) \nonumber \\
    &\gtrapprox \hat{\Entr}(C \mid G) - \hat{\Entr}_q(C \mid W, G). \nonumber
\end{align*}
We note that these are expected lower bounds, i.e. they are \emph{exact} when taking
an expectation under the true distribution $p$.
We cannot make a similar statement about tripartite MI, though, since it is computed as the difference of two lower-bound approximations of true mutual information quantities. 

\section{Results}\label{results}
Our main experimental results are presented in \autoref{tab:results-form}. We find that both form and lexical semantics significantly interact with declension class in both Czech and German (each $p < 0.01$).\footnote{All results in this section were significant for both languages, according to a \citet{welch1947generalization}'s $t$-test, which yielded $p<0.01$ after \citeauthor{bhCorrection}'s correction. A \citet{welch1947generalization}'s $t$-test differs from \citet{student1908}'s $t$-test in that the latter assumes equal variances, and the former does not, making it preferable (see \citealt{delacre2017}).} We observe that our estimates of $\MI(C; W \mid G)$ is larger (0.5--0.8 bits) than our estimates of $\MI(C; V \mid G)$ (0.3--0.5 bits). We also observe that the $\MI$ estimates in Czech are higher than in German. However, 
we caution that the unnormalized estimates for the two languages are not fully comparable because they hail from models trained on different amounts of data.
The tripartite $\MI$ estimates between class, form, and meaning, were relatively small (0.2--0.35 bits) for both languages. We interpret this finding as showing that much of the information contributed by form is not redundant with information contributed by meaning---although a substantial amount is.

\begin{table}
\begin{center}
\small
\begin{adjustbox}{width=\linewidth}
\begin{tabular}{lcccc}
\toprule
 & $\Entr(C)$ & $\Entr(C \mid G)$ & $\MI(C ; G)$ & $\U(C \mid G)$  \\ \midrule
Czech  & 2.75 & 1.35 & \bf 1.40 & 50.8\% \\ 
German & 2.88 & 2.17 & \bf 0.71 & 24.6\% \\ \bottomrule
\end{tabular}
\end{adjustbox}
\end{center}
 \caption{MI between class and gender $\MI(C ; G)$: $\Entr(C)$ is class entropy, $\Entr(C \mid G)$ is class entropy given gender, $U(C \mid G)$ is  the uncertainty coefficient. 
 } \label{tab:results-gender}
\end{table}
 
\begin{figure*}[t]
\begin{subfigure}[t]{0.46\textwidth}
    \centering
    \includegraphics[width=\columnwidth]{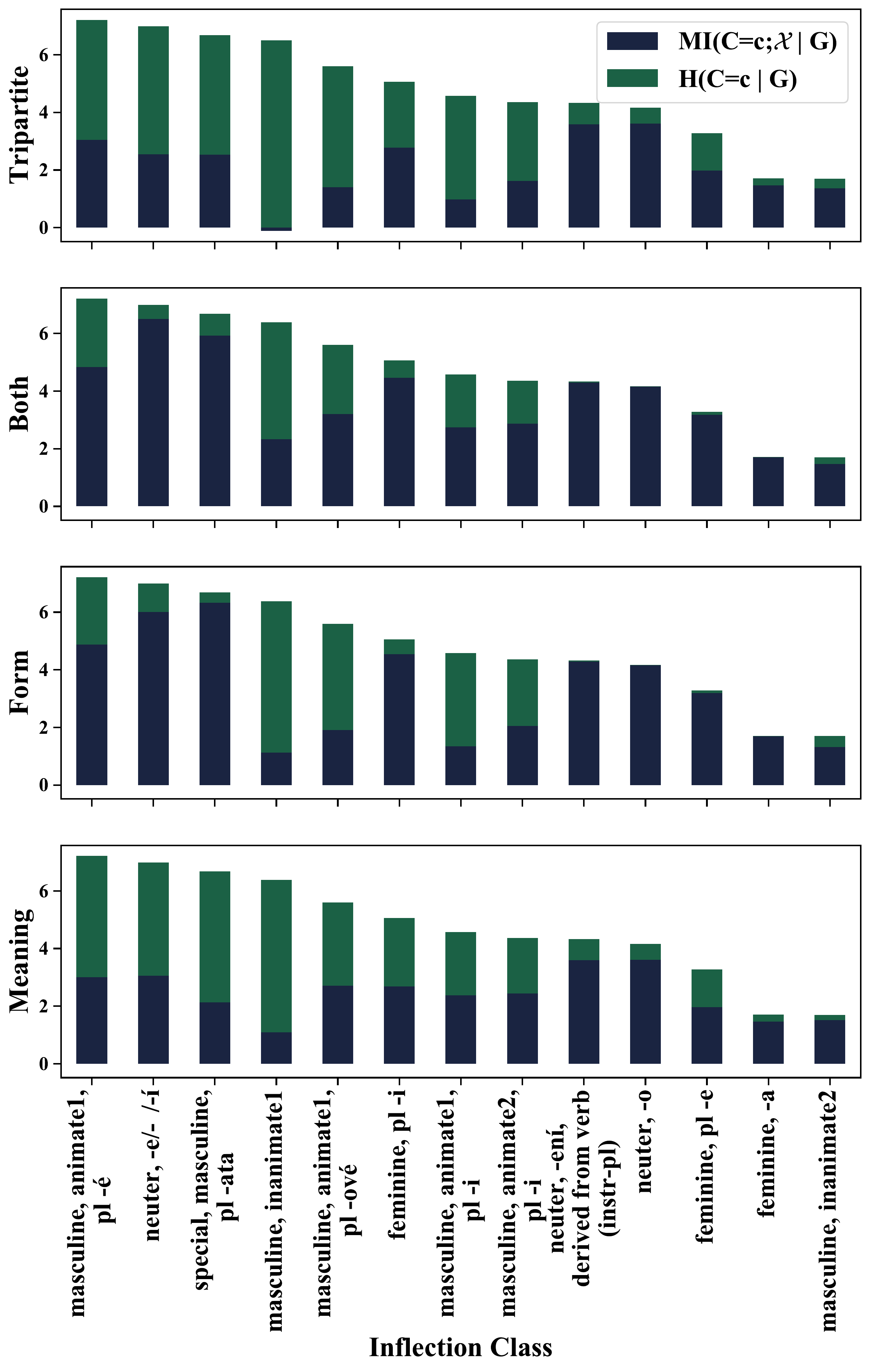}
    \caption{Czech}\label{fig:CzechNouns_new}
\end{subfigure}%
\hfill
\begin{subfigure}[t]{0.48\textwidth}
    \centering
    \includegraphics[width=\columnwidth]{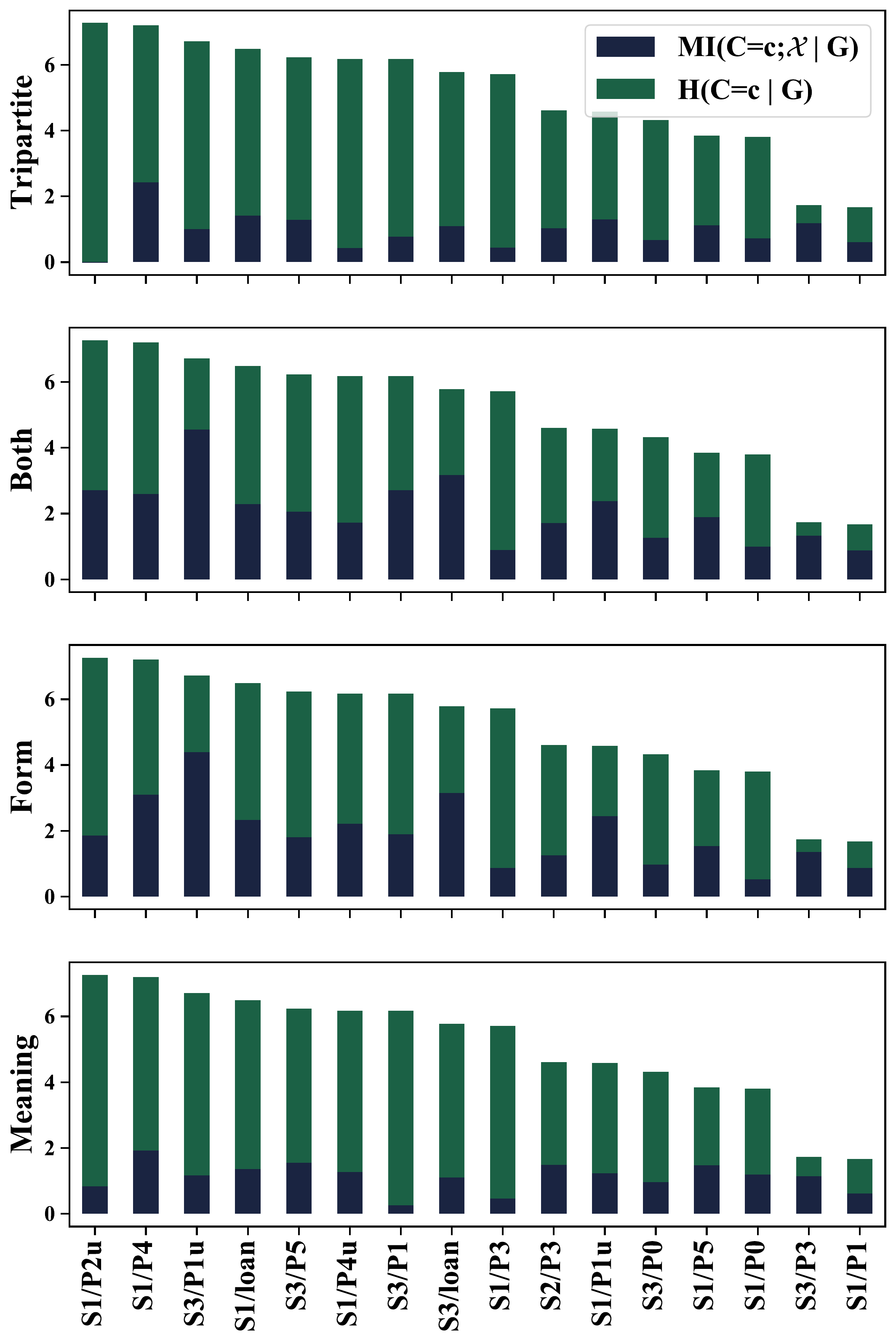}
    \caption{German}\label{fig:GermanNouns_new}
\end{subfigure}
\caption{Pointwise $\MI$ for declension classes. $\MI$ for each random variable $\mathcal{X} \in \{ V, W, \{V,W\}\ , \{V;W\}\}$ are plotted for classes increasing in size (towards the right): $\MI(C = c; V | G)$ (bottom), $\MI(C = c; W | G)$ (bottom middle), $\MI(C =c ; V, W | G)$ (top middle), and  tripartite $\MI(C = c; V; W | G)$ (top).}\label{fig:Nouns_new}
\end{figure*}%

As a final sanity check, we measure mutual information between class and gender $\MI(C; G)$ (see \autoref{tab:results-gender}). For both languages, the mutual information between declension class and gender is significant. Our $\MI$ estimates range from approximately $\sfrac{3}{4}$ of a bit in German up to 1.4 bits in Czech, which respectively amount to nearly 25\% and nearly 51\% of the remaining unconditional entropy. Like the quantities discussed in \autoref{sec:methods}, this $\MI$ was estimated using simple plug-in estimation. Remember, if class were entirely reducible to gender, conditional entropy of class given gender would be zero. This is not the case: Although the conditional entropy of class given gender is lower for Czech (1.35 bits) than for German (2.17 bits), in neither case is declension class informationally equivalent to the language's grammatical gender system. 

\section{Discussion and Analysis}

Next, we ask whether individual declension classes \emph{differ} in how idiosyncratic they are, e.g.,\ does any one German declension class share less information with form than the others? To address this, we qualitatively inspect per-class half-pointwise mutual information in \autoref{fig:CzechNouns_new}--\ref{fig:GermanNouns_new}. See \autoref{tab:results5-meaning} in the Appendix for the five highest and lowest surprisal examples per model. 
Several qualitative trends were observed:\ (i) classes show a decent amount of variability, (ii) unconditional entropy for each class is inversely proportional to the class' size, (iii) half-pointwise $\MI$ is higher on average for Czech than German, and (iv)  classes that have high $\MI(C=c;V \mid G)$ usually have high $\MI(C=c;W \mid G)$ (with a few notable exceptions we discuss below). 
\paragraph{Czech.} 

In general, declension classes associated with masculine nouns ($g=\textsc{msc}$) have smaller $\MI(C=c ; W\mid G)$ than classes associated with feminine ($g=\textsc{fem}$) and neuter ($g=\textsc{neu}$) ones of a comparable size---the exception being `special, masculine, plural -\textit{ata}'. This class ends exclusively in \textit{-e} or \textit{-\u{e}}, which might contribute to that class' higher $\MI(C=c; W \mid G)$. That $\MI(C=c ; W \mid G)$ is high for feminine and neuter classes suggests that the overall $\MI(C ; W \mid G)$ results might be largely driven by these classes, which predominantly end in vowels. We also note that the high $\MI(C=c ; W \mid G)$ for feminine `plural \textit{-e}', might be driven by the many Latin or Greek loanwords present in this class.

With respect to meaning, masculine declension classes can reflect degrees of animacy: `animate1' contains nouns referring mostly to humans and a few animals (\textit{kocour} `tomcat', \textit{\u{c}olek} `newt'), `animate2' contains nouns referring mostly to animals and a few humans (\textit{syn} `son', \textit{k\v{r}est'an} `Christian'), `inanimate1' contains many plants, staple foods (\textit{chl\'eb} `bread', \textit{ocet} `vinegar') and meaningful places (\textit{domov} `home', \textit{kostel} `church'), and `inanimate2' contains many basic inanimate nouns (\textit{k\'amen} `stone'). Of these masculine classes, `inanimate1' has a lower $\MI(C=c ; V \mid G)$ than its class size alone might lead us to predict. Feminine and neuter classes show no clear pattern, although neuter classes `\textit{-eni}' and `\textit{-o}' have comparatively high $\MI(C=c ; V \mid G)$.

For $\MI(C=c ; V; W \mid G)$, we observe that `masculine, inanimate1' is the smallest quantity, followed by most other masculine classes (e.g., masculine animate classes with \textit{-ov\'e} or \textit{-i} plurals) for which $\MI(C=c; W \mid G)$ was also low. Among non-masculine classes, we observe that feminine `pl -\textit{i}' and the neuter classes \textit{-o}  and \textit{-en\'i} show higher tripartite MI. The latter two classes have relatively high $\MI$ across the board.

\paragraph{German.} $\MI(C=c ; W \mid G)$ for classes containing words with umlautable vowels (i.e., S3/P1u, S1/P1u) or loan words (i.e., S3/loan) tends to be high; in the prior case, our models seem able to separate umlautable from non-umlautable vowels, and in the latter case, loan word orthography from native orthography. $\MI(C=c ; V \mid G)$ quantities are roughly equivalent across classes of different size, with the exception of three classes: S1/P4, S3/P1, and S1/P3. S1/P4 consists of highly semantically variable nouns, ranging from relational noun lexemes (e.g., \textit{Glied} `member', \textit{Weib} `wife', \textit{Bild} `picture') to masses (e.g., \textit{Reis} `rice'), which perhaps explains its relatively high $\MI(C=c ; V \mid G)$. For S1/P3 and S3/P1, $\MI(C=c ; V \mid G)$ is low, and we observe that both declension classes idiosyncratically group clusters of semantically similar nouns: S1/P3 contains ``exotic'' birds (\textit{Papagei} `parrot', \textit{Pfau} `peacock'), but also nouns ending in \textit{-or}, (\textit{Trakt\textbf{or}} `tractor', \textit{Past\textbf{or}} `pastor'), whereas S3/P1 contains very few nouns, such as names of months (\textit{M\"arz}, `March', \textit{Mai} `May') and names of mythological beasts (e.g., \textit{Sphinx}, \textit{Alp}).  
 
Tripartite $\MI$ is fairly idiosyncratic in German: The lowest quantity comes from the smallest class, S1/P2u. S1/P3, a class with low $\MI(C=c ; V \mid G)$ from above, also has low tripartite $\MI$. We speculate that S1/P3 could be a sort of ``catch-all'' class with no clear regularities. 
The highest tripartite $\MI$ comes from S1/P4, which also had high $\MI(C=c ; V \mid G)$. The existence of significant tripartite $\MI$ results suggests that submorphemic meaning bearing units, or phonaesthemes, might be present. Taking inspiration from \citealt{pimentel2019}, which aims to automatically discover such units, we observe that many words in S1/P4 contain letters \{\textit{d, e, g, i, l}\}, often in identically ordered orthographic sequences, such as \textit{B\textbf{ild}, B\textbf{ie}st,
F\textbf{eld}, \textbf{Geld}, \textbf{Glied},  K\textbf{i}n\textbf{d}, 
\textbf{Lei}b, \textbf{Lied}, Sch\textbf{ild}, V\textbf{ie}ch, W\textbf{ei}b}, etc. While these letters are common in German orthography, their noticeable presence suggests that further elucidation of declension classes in the context of phonaesthemes could be warranted.

\section{Conclusion}
We adduce new evidence that declension class membership is not wholly idiosyncratic nor fully deterministic based on form or meaning in Czech and German. We quantify mutual information and find estimates which range from 0.2 bits to nearly one bit. Despite their relatively small magnitudes, our estimates of mutual information between class and form accounted for between 25\% and 60\% of the class' entropy, even after relevant controls, and $\MI$ between class and meaning accounted for between 13\% and nearly 40\%. We analyze results per-class, and find that classes vary in how much information they share with meaning and form. We also observe that classes that have high $\MI(C=c ; V \mid G)$ often have high $\MI(C=c ; W \mid G)$, with a few noted exceptions that have specific orthographic (e.g., German umlauted plurals), or semantic (e.g., Czech masculine animacy) properties. In sum, this paper has proposed a new information-theoretic method for quantifying the strength of morphological relationships, and applied it to declension class. We verify and build on existing linguistic findings, by showing that the mutual information quantities between declension class, orthographic form, and lexical semantics are statistically significant. 

\section*{Acknowledgments}

Thanks to Guy Tabachnik for discussions on Czech phonology, to Jacob Eisenstein for useful questions about irregularity, and to Andrea Sims and Jeff Parker for advice on citation forms. Thanks to Ana Paula Seraphim for helping beautify \autoref{fig:declension_mi}. 

\bibliographystyle{acl_natbib}
\bibliography{acl2019}
\clearpage

\appendix

\begin{table*}[h!]
\begin{center}
\small
    \begin{tabular}{lrllrll}
\toprule
    \multicolumn{3}{c}{\bf Czech} 
    & \multicolumn{4}{c}{\bf German}\\ 
    \cmidrule(l{.5em}){1-3} \cmidrule(l{.5em}){4-7}
class & \#  & gender & class & \#  & classic class & gender(s) \\ \midrule
 masculine, inanimate2    &           823 & \textsc{msc}  & S1/P1    &     1157    &      Decl I    & \textsc{msc}, \textsc{neut} \\
 feminine, \textit{-a}  &     818 & \textsc{fem}& S3/P3    &    1105     &     Decl VI & \textsc{fem}  \\
 feminine, pl \textit{-e}   &   275 & \textsc{fem} & S1/P0    &     264    &      Singularia Tantum & \textsc{msc}, \textsc{neut}, \textsc{fem} \\
  neuter, \textit{-o} &  149 & \textsc{neut} & S1/P5    &     256    &      ``default \textit{-s} PL'' & \textsc{msc}, \textsc{neut}, \textsc{fem} \\
 neuter, \textit{-en\'i}  &  133 & \textsc{neut} & S3/P0    &    184     &     Singularia Tantum & \textsc{msc}, \textsc{neut}, \textsc{fem} \\
 masculine, animate2, pl \textit{-i})  &  130 & \textsc{msc}& S1/P1u    &   154     &     Decl II & \textsc{msc} \\
 masculine, animate1, pl \textit{-i})  &  112 & \textsc{msc}& S2/P3     &    151      &    Decl V & \textsc{msc} \\
 feminine, pl \textit{-i}    &   80 & \textsc{fem}& S1/P3     &    70   &       Decl IV & \textsc{msc}, \textsc{neut}  \\
  masculine, animate1, pl \textit{-ov\'e}  &   55 & \textsc{msc}& S3/loan   & 67 &  Loanwords & \textsc{msc}, \textsc{neut}, \textsc{fem} \\
 masculine, inanimate1  &   32 & \textsc{msc} & S3/P1  &       51    &      Decl VIII & \textsc{fem}    \\
 special, masculine, pl \textit{-ata}  &   26 & \textsc{msc}& S1/P4u    &    51   &       Decl III & \textsc{msc}, \textsc{neut}  \\
  neuter, \textit{-e/-\u{e}/-\'i}    &   21 & \textsc{neut} & S3/P5    &     49    &      ``default \textit{-s} PL'' & \textsc{msc}, \textsc{neut}, \textsc{fem} \\
 masculine, animate1, pl \textit{-\'e}  &   18 & \textsc{msc} & S1/loan    &    41    &      Loanwords & \textsc{msc}, \textsc{neut} \\
   &       &   & S3/P1u    &    35    &      Decl VII & \textsc{fem} \\
 &   &   & S1/P4     &    25      &    Decl III & \textsc{msc}, \textsc{neut} \\
   &       &   & S1/P2u   &   24      &    Decl II & \textsc{msc}, phon. \\
 \midrule
Total & 2672 & & & 3684 & &  \\\bottomrule
    \end{tabular}
\end{center}
    \caption{Declension classes.\ `class' refers to the declension class identifier,  `\#' refers to the number of lexemes in each declension class, and `gender' refers to the gender(s) present in each class. German declension classes came from CELEX2, for which `S' refers to a noun's singular form, `P' refers to its plural, `classic class' refers to the conception of class from \textit{Brockhaus Wahrig W{\"o}rterbuch}.}
    \label{tab:NounClass}
\end{table*}

 \section{Further Notes on Preprocessing}\label{append:classes}

The breakdown of our declension classes is given in \autoref{tab:NounClass}. We will first discuss more details about our preprocessing and linguistic analysis for Czech, and then for German.

\paragraph{Czech.} The Czech classes were initially derived from an edit-distance heuristic between nouns. A fluent speaker--linguist then identified major noun classes by grouping together nouns with shared suffixes in the surface (orthographic) form. If the differences between two sets of suffixes in the surface form could then be accounted for by positing a basic phonological rule---for example, vowel shortening in monosyllabic words---then the two sets were collapsed. 

\begin{figure}[ht]
    \centering
    \includegraphics[width=\columnwidth]{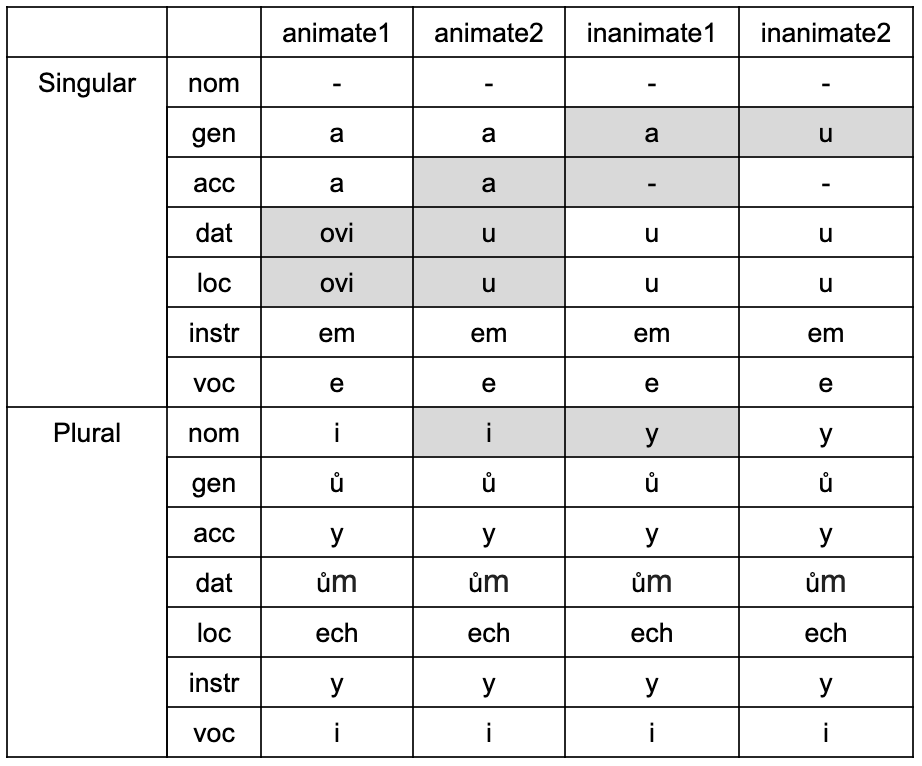}
    \caption{Czech paradigm for masculine nouns.}
    \label{fig:czechparad}
\end{figure}

Among masculine nouns, four large classes were identified that seemed to range from ``very animate'' to ``very inanimate.'' The morphological divisions between these classes were very systematic, but there was substantial overlap: dat.sg and loc.sg differentiated `animate1' from `animate2', `inanimate1' and `inanimate2'; acc.sg, nom.pl and voc.pl differentiated `animate2' from `inanimate1' and `inanimate2', and gen.sg differentiated `inanimate1' from `inanimate2' (see \autoref{fig:czechparad}). Further subdivisions were made within the two animate classes for the apparent idiosyncratic nominative plural suffix, and within the `inanimate2' class, where nouns took either \textit{-u} or \textit{-e} as the genitive singular suffix. This division may have once reflected a final palatal on nouns taking \textit{-e} in the genitive singular case, but this distinction has since been lost. All nouns in the `inanimate2' ``soft'' class end in coronal consonants, whereas nouns in the `inanimate1' ``hard'' class have a variety of final consonants.

Among feminine nouns, the `feminine -a' class contained all feminine words that ended in \textit{-a} in the nominative singular form. (Note that there exist masculine nouns ending in \textit{-a}, but these did not pattern with the `feminine -a' class). The `feminine pl -e' class contained feminine nouns ending in \textit{-e}, \textit{-\v{e}}, or a consonant, and as the name suggests, had the suffix  \textit{-e} in the nominative plural form. The `feminine pl -i' class contained feminine nouns ending in a consonant and had the suffix  \textit{-i} in the nominative plural form. No feminine nouns ended in a dorsal consonant. 

Among neuter nouns, all words ended in a vowel. 

\paragraph{German.} After extracting declension classes from CELEX2, we made some additional preprocessing decisions for German, usually based on orthographic or other considerations. For example, we combined the classes S1 with S4 classes, P1 with P7, and P6 with P3 because the difference between each member of any of these pairs lies solely in spelling (a final $<$s$>$ is doubled in the spelling when GEN.SG \textit{-(e)s}, or the PL \textit{-(e)n} is attached).

Whether a given singular, say S1, becomes inflected as P1 or P2---or, for that matter, the corresponding umlauted versions of these plural classes---is phonologically conditioned \citep{alexiadou2008}. If the stem ends in a trochee whose second syllable consists of schwa plus /n/, /l/, or /r/, the schwa is not realized, i.e., it gets P2, otherwise it gets P1. For this phonological reason, we also chose to collapse P1 and P2.

We also collapsed all loan classes (i.e., those with P8--P10) under one plural class `Loan'. This choice resulted in us merging loans with Greek plurals (like P9, \textit{Myth-os / \textbf{Myth-en}}) with those with Latin plurals (like P8, \textit{Maxim-um / \textbf{Maxim-a}} and P10, \textit{Trauma / \textbf{Trauma-ta}}). This choice might have unintended consequences on the results, as the orthography of Latin and Greek differ substantially from each other, as well as from the native German orthography, and might be affecting our measure of higher form-based $\MI$ for S1/Loan and S3/Loan classes in Table 3 of the main text. One could reasonably make a different choice, and instead remove these examples from consideration, as we did for classes with fewer than 20 lemmata.

\section{Some prototypical examples}

\begin{table*}[h]
\begin{center}
\small
\begin{tabular}{lcclcc}
\toprule
 \multicolumn{3}{c}{\bf Czech} & \multicolumn{3}{c}{\bf German} \\ 
  stem & class & $\Entr(C \mid W, V)$ & stem & class & $\Entr(C \mid W, V)$ \\
 \midrule
 \it pavouk & masculine, animate2, pl \textit{-i} & 11.54 & \it Balance & \textsc{fem}, ?, S1P5 & 13.16 \\
 \it investor & masculine, animate2, pl \textit{-i} & 10.93 & \it Hertz & \textsc{neut}, ?, S3P0 & 13.05 \\
 \it v\r{u}l & masculine, animate2, pl \textit{-i} & 10.78 & \it Schmack & \textsc{msc}, 6, S3P3 & 12.17 \\
 \it dla\u{z}di\u{c} & masculine, animate1, pl \textit{-ov\'e} & 10.01 & \it See & \textsc{fem}, 6, S3P3 & 12.12 \\ 
  \it opylova\u{c} & masculine, animate2, pl \textit{-i} & 9.21 & \it Reling & \textsc{fem}, ?, S3P5 & 11.81 \\ \hline
 \it optika & feminine, \textit{-a} & $2.2 x 10^{-4}$ & \it Glocke & \textsc{fem}, 6, S3P3 & $5.7 x 10^{-3}$ \\
 \it kritika & feminine, \textit{-a} & $2.2 x 10^{-4}$ & \it Schale & \textsc{fem}, 6, S3P3 & $5.7 x 10^{-3}$ \\
 \it pahorkatina & feminine, \textit{-a} & $2.1 x 10^{-4}$ & \it Schnecke & \textsc{fem}, 6, S3P3 & $5.6 x 10^{-3}$ \\
 \it kachna & feminine, \textit{-a} & $2.1 x 10^{-4}$ & \it Zeche & \textsc{fem}, 6, S3P3 & $5.6 x 10^{-3}$ \\
 \it matematika & feminine, \textit{-a} & $2.1 x 10^{-4}$ & \it Parzelle & \textsc{fem}, 6, S3P3 & $4.8 x 10^{-3}$ \\
\bottomrule
\end{tabular}
\end{center}
\caption{Five highest and lowest surprisal examples given form and meaning (w2v) by language.} \label{tab:results5-meaning}
\end{table*}


To explore which examples, across classes might be most prototypical, we sampled the top five highest and lowest suprisal examples. The results are in \autoref{tab:results5-meaning}. We observe that the lowest surprisal forms for each language generally come from a single class for each language:  feminine, \textit{-a} for Czech and S3/P3 for German. These two classes were among the largest, having lower class entropy, and both contained feminine nouns. Forms with higher surprisal generally came from several smaller classes, and were predominately masculine. This sample size is small however, so it remains to be investigated whether this tendency in our data belies a genuine statistically significant relationship between gender, class size, and surprisal.
\end{document}